\documentclass[letterpaper, 10 pt, conference]{ieeeconf} 

\IEEEoverridecommandlockouts
\usepackage{cite}
\usepackage{amsmath,amssymb,amsfonts}
\usepackage{algorithm}
\usepackage{algpseudocode}
\usepackage{graphicx}
\usepackage{textcomp}

\usepackage{hyperref}

\usepackage{xcolor}
\definecolor{dark-red}{rgb}{0.4,0.15,0.15}
\definecolor{dark-blue}{rgb}{0.15,0.15,0.8}
\definecolor{medium-blue}{rgb}{0,0,0.5}
\hypersetup{
    colorlinks, linkcolor={dark-red},
    citecolor={dark-blue}, urlcolor={medium-blue}
}

\usepackage{tikz}

\newcommand\copyrighttext{%
  \footnotesize  \textcopyright Accepted at 2023 IEEE/International Conference on Robotics and Automation (ICRA)}
\newcommand\copyrightnotice{%
\begin{tikzpicture} [remember picture,overlay]
 \node[anchor=south,yshift=10pt] at (current page.south)
    {\copyrighttext};
\end{tikzpicture}%
}

\usepackage{subcaption} 
\usepackage{siunitx} 
\usepackage[T1]{fontenc}

\begin{document}
%
%
%
\definecolor{newChanges}{RGB}{0,0,0}
\definecolor{needChanges}{RGB}{127,0,255}
\definecolor{copiedFromMRS}{RGB}{0,0,255}
\definecolor{question}{RGB}{255,0,0}
\definecolor{ToDo}{RGB}{153, 51, 0}

%
%
%
%
\title{\LARGE \bf
Estimation of continuous environments by robot swarms:\\
Correlated networks and decision-making*
}

\author{Mohsen Raoufi$^{1,2,3}$, Pawel Romanczuk$^{1,2,4}$ and Heiko Hamann$^{1,5}$ %
\thanks{*This work is funded by the Deutsche Forschungsgemeinschaft (DFG, German Research Foundation) under Germany’s Excellence Strategy – EXC 2002/1 “Science of Intelligence” – project number 390523135.}
\thanks{$^{1}$ Mohsen Raoufi, Pawel Romanczuk and Heiko Hamann are with Science of Intelligence, Research Cluster of Excellence, Marchstr. 23, 10587 Berlin, Germany}
\thanks{$^{2}$ Mohsen Raoufi and Pawel Romanczuk are with Institute for Theoretical Biology, Department of Biology, Humboldt Universität zu Berlin, Berlin, Germany {\tt\small mohsen.raoufi@hu-berlin.de}, {\tt\small pawel.romanczuk@hu-berlin.de} }
\thanks{$^{3}$ Mohsen Raoufi is with Department of Electrical Engineering and Computer Science, Technical University of Berlin, Berlin, Germany}
\thanks{$^{4}$ Pawel Romanczuk is with Bernstein Center for Computational Neuroscience, Berlin, Germany}
\thanks{$^{5}$ Heiko Hamann is with Department of Computer and Information Science, University of Konstanz, Konstanz, Germany {\tt\small  hamann@uni-konstanz.de}}
}
\maketitle
%
%
%
%
\begin{abstract}
Collective decision-making is an essential capability of large-scale multi-robot systems to establish autonomy on the swarm level. 
A~large portion of literature on collective decision-making in swarm robotics focuses on discrete decisions selecting from a limited number of options. 
Here we assign a decentralized robot system with the task of exploring an unbounded environment, finding consensus on the mean of a measurable environmental feature, and aggregating at areas where that value is measured (e.g., a~contour line). 
A~unique quality of this task is a causal loop between the robots' dynamic network topology and their decision-making. 
For example, the network's mean node degree influences time to convergence while the currently agreed-on mean value influences the swarm's aggregation location, hence, also the network structure as well as the precision error. 
We propose a control algorithm and study it in real-world robot swarm experiments in different environments. 
We show that our approach is effective and achieves higher precision than a control experiment. 
We anticipate applications, for example, in containing pollution with surface vehicles. 
\end{abstract}
%
%
%
%
\section{Introduction}
\copyrightnotice
Collective decision-making in large-scale decentralized multi-robot systems is required to coordinate and organize the system~\cite{raoufi2021speed, ebert2020bayes, brambilla2013swarm, valentini2016collective}. For example, a robot swarm needs to collectively agree on a common direction in flocking or on a task allocation~\cite{raoufi2019self}. 
While task allocation is an example for a discrete consensus problem similar to best-of-$n$ problems (collectively choosing from a finite and countable set), the flocking example is a continuous consensus achievement problem~\cite{valentini2017best}.
Large portions of the collective decision-making literature in swarm robotics are focused on discrete problems, such as the popular collective perception benchmark scenario~\cite{valentini2016collective}. 
Here we focus on a continuous consensus achievement problem ~\cite{olfati2007consensus, ding2022consensus} in the form of a decentralized estimation scenario~\cite{leonard2022collective}. In our previous work we studied the effect of diverse information on the accuracy of collective estimation, which forms the exploration-exploitation trade-off~\cite{raoufi2021speed}. To achieve diverse-enough information, the swarm needs to expand and sample from larger area, which leads to a dispersal collective behavior. Among the proposed distributed methods in the literature on dispersion, some use information that is either costly or not available for all swarm platforms~\cite{ugur2007dispersion}. However, an approximate estimation of distance proved to be efficient 
to achieve such a goal. The performance of greedy gradient descent algorithm for dispersion predicted to be challenging, especially with large number of robots ($N>10$)~\cite{bayert2019robotic}. Thus, to overcome this, we propose a threshold-based random walk algorithm that proves to be efficient {\color{newChanges}enough} for larger swarms ($N=40$).
\par
In addition, we require a form of exploitation of the collective decision as the robots need to react to their collective decisions and aggregate at areas that are determined by their consensus. 
This comes with a design challenge. Should the robots separate a consensus finding phase from an exploitation phase? Either they synchronize and determine an end of the collective decision-making process or they asynchronously switch to exploitation and try to keep finding a consensus on the go. 
Here we propose a solution choosing the asynchronous option. 
Consequently, we face another challenge. As the robots initiate their exploitation process, they try to move towards the designated area while continuing to communicate with neighbors. They form a dynamic network topology while following the collective decision-making protocol. We know that the network topology influences the decision-making process~{\color{newChanges}\cite{srivastava2014collective, lobel2016preferences, becker2017network, mateo2019optimal, kwa2023effect}} and hence the emerging process is self-referential (network influences consensus, consensus influences spatial displacement). In that regards, there is a huge body of literature studying this effect from a network point of view. An example of such phenomenon is the homophily in social networks~\cite{khanam2022homophily, holme2006nonequilibrium}. However, studying the co-evolution of network and opinion dynamics in swarm robotics has been overlooked. In this paper, we show how the swarm of real robots disperse in an unbounded environment and then aggregating at the points where they agreed on. 
\par
\newcommand\figFourWidth{1.9}
\begin{figure*}
\centering
    \subcaptionbox{Initial Distribution}{\includegraphics[height=\figFourWidth in]{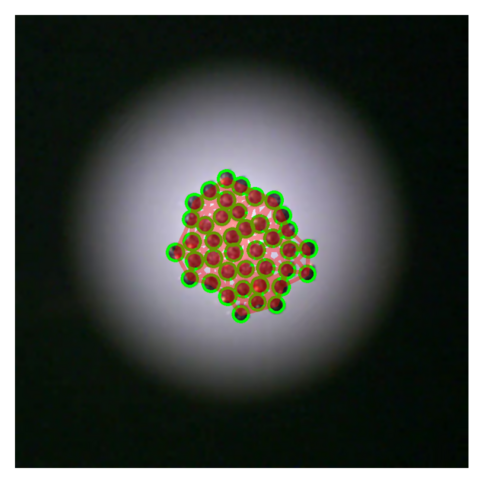}}%
    \hfill
    \subcaptionbox{Dispersed}{\includegraphics[height=\figFourWidth in]{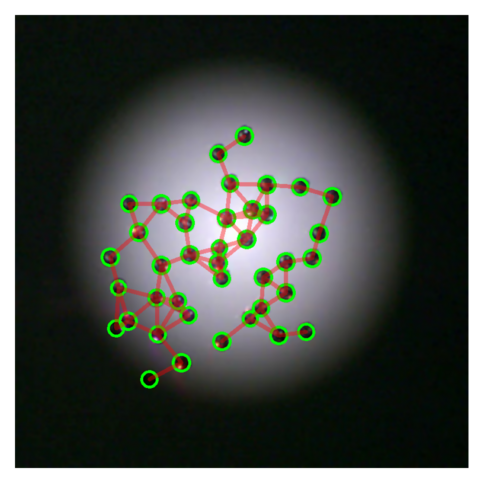}}%
    \hfill
    \subcaptionbox{Final Consensus}{\includegraphics[height=\figFourWidth in]{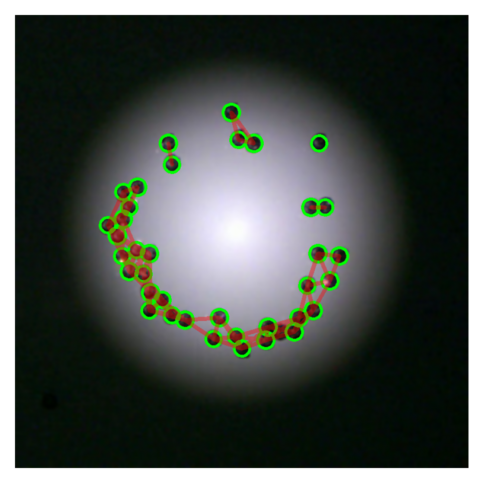}}%
    \hfill
    \subcaptionbox{Kilobot with light conductor}{\includegraphics[height= \figFourWidth in]{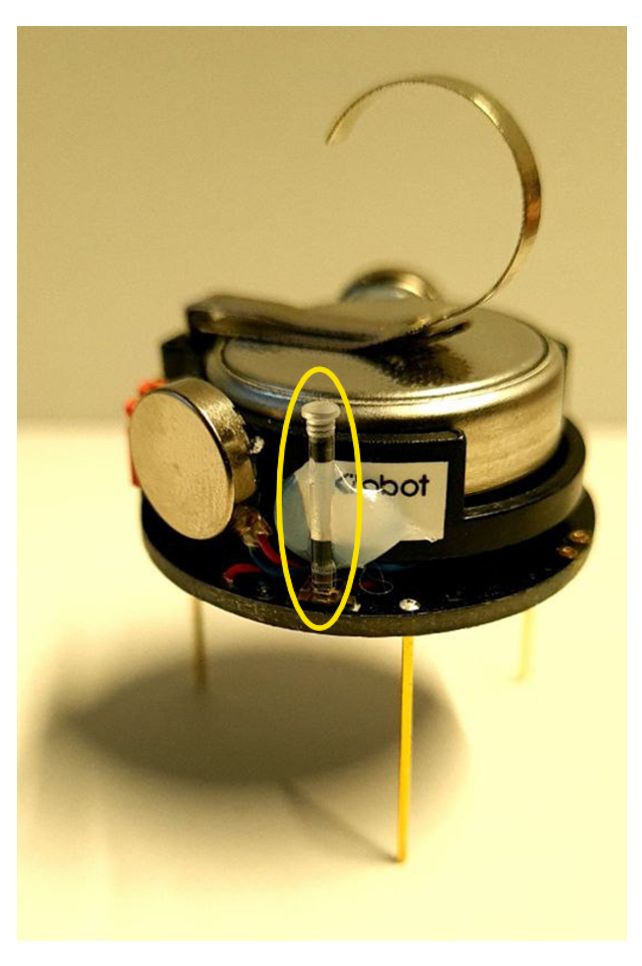}}%
\caption{a-c) Snapshots from the top-view camera with detected real Kilobots (green circles) in a radial (cone-shaped) light distribution; the red lines show the possible link between two robots within the communication range. d)~The light conductor (transparent plastic) added to the Kilobots to solve the issue of shadows that are cast from robot bodies on their light sensor.}
\label{fig:snapShots_pl_ATP_err_XprXpt}
\end{figure*}
%
%
%
%
%
\section{Method}
Following our previous work~\cite{raoufi2021speed}, we study the co-evolution of network structure and collective estimation for a swarm of $N$ real robots. The value to estimate is a continuous, spatially distributed scalar feature of the environment. In our experiments, this will be realized by a spatially varying light intensity field. The swarm's goal is to estimate a global property of the distributed feature and approach it in the physical space. Our focus is on estimation and localization of the environmental field's mean value {\color{newChanges}(see Fig.~\ref{fig:snapShots_pl_ATP_err_XprXpt})}. 
\par
We define two phases: exploration and exploitation. Having separate phases for exploration and exploitation has been shown to be more efficient than mixed phases~\cite{reina2015design}. During initial exploration (see Sec.~\ref{subsec:exploration}), we program the swarm to expand. The aim is for the individual robots to collect diverse estimates of the environmental feature. The robots are supposed to cover as much area as possible while keeping network largely connected. 
{\color{newChanges} The communication range and the swarm size determine how much the swarm can expand without being disconnected. We define the end of exploration as the moment when the collective achieves a maximal area coverage while still maintaining connectivity.} During the subsequent exploitation phase (see Sec.~\ref{subsect:exploitation}), robots communicate to achieve a consensus on the mean value, and at the same time, try to move toward the spots in the environment where the measured intensity is closer to the consensus.
We showed previously that by combining these components a contour-capturing behavior emerges~\cite{raoufi2021speed}. A~possible application is to contain pollution or localize the position of a resource in the environment~\cite{zahugi2012design, kaviri2019coverage, amjadi2019cooperative, haghighat2022approach}. 
\par
\newcommand\figTwoWidth{1.6}
\begin{figure}[!b]%
\centering
\includegraphics[width=0.7\linewidth]{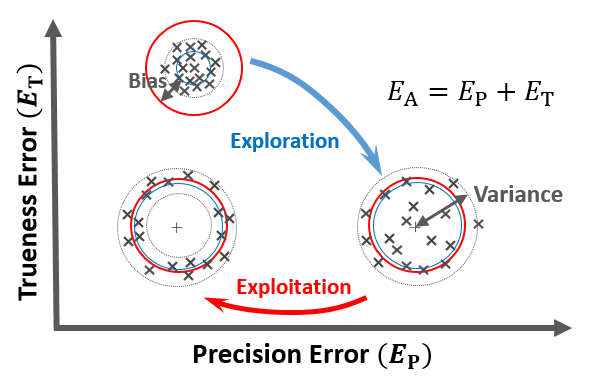}  
\caption{The relation between accuracy (trueness and precision) errors in collective contour capturing for the example of radial distribution. The red and blue circles show the ground truth and collective mean contours respectively, and the crosses are the positions of robots in physical space. The initial trueness error (top left circle) is reduced during exploration phase, whereas the precision error increases (bottom right). During exploration, the precision error decreases~\cite{raoufi2021speed}, and robots capture the contour (bottom left).} 
\label{fig:ATP_err__XprXpt}
\end{figure}
We minimized the requirements with respect to the robotic platform to enable the implementation of the algorithm even on minimal robots, here specifically the  Kilobot platform~\cite{rubenstein2012kilobot}. Although some algorithmic details are specific to our implementation on Kilobots, our model is generally applicable regardless of the swarm robotic platform. The requirements are: \textbf{a)} fully distributed algorithm; no central control, \textbf{b)} only local environmental information available, \textbf{c)} communication only to local neighbors, within a limited communication range, \textbf{d)} no prior information neither about the environment, nor the neighbors, and \textbf{e)} unbounded arena.
\par
\renewcommand\figTwoWidth{1.3}
\begin{figure*}[!t]
\centering
    \subcaptionbox{}{\includegraphics[height=\figTwoWidth in]{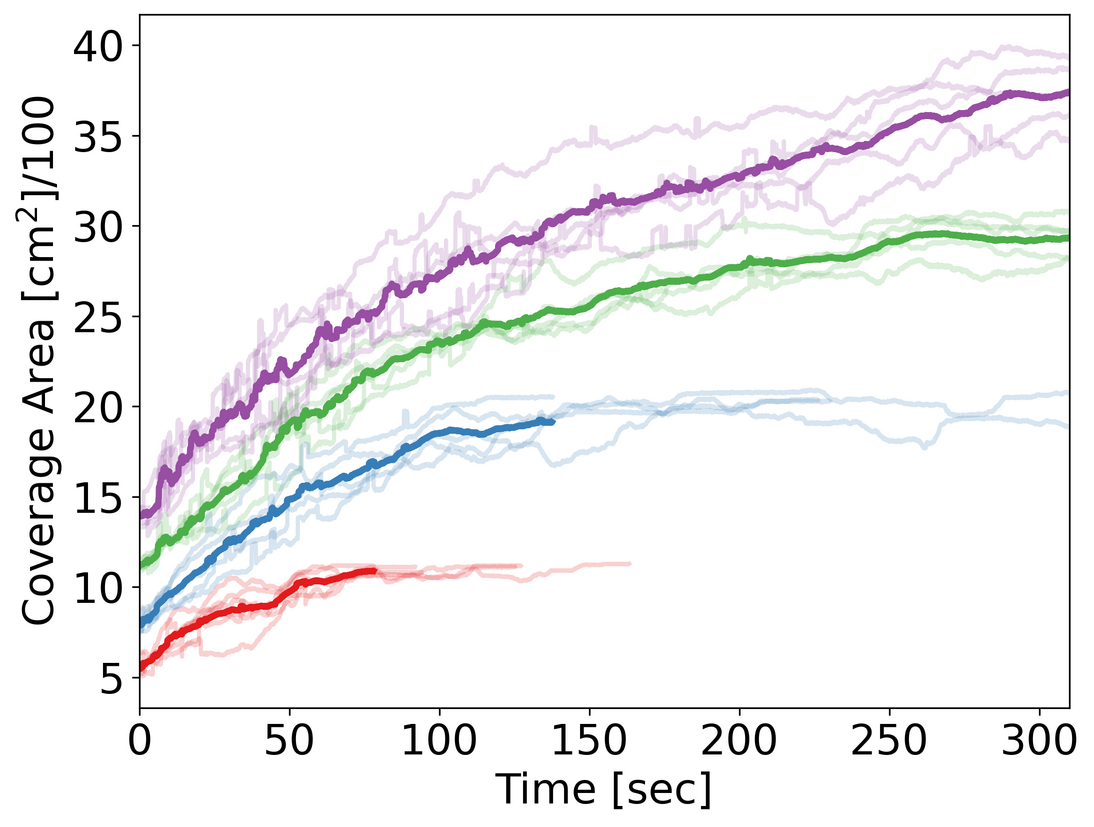}}%
    \hfill
    \subcaptionbox{}{\includegraphics[height=\figTwoWidth in]{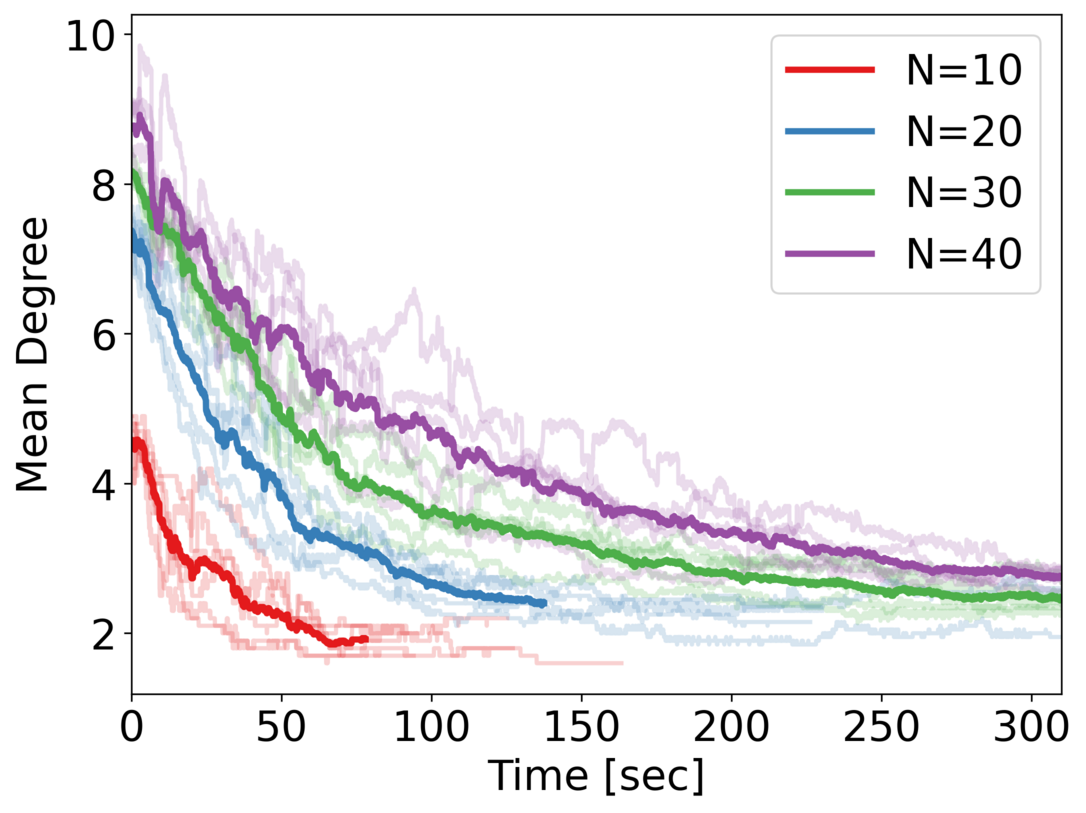}}%
    \hfill
    \subcaptionbox{}{\includegraphics[height=\figTwoWidth in]{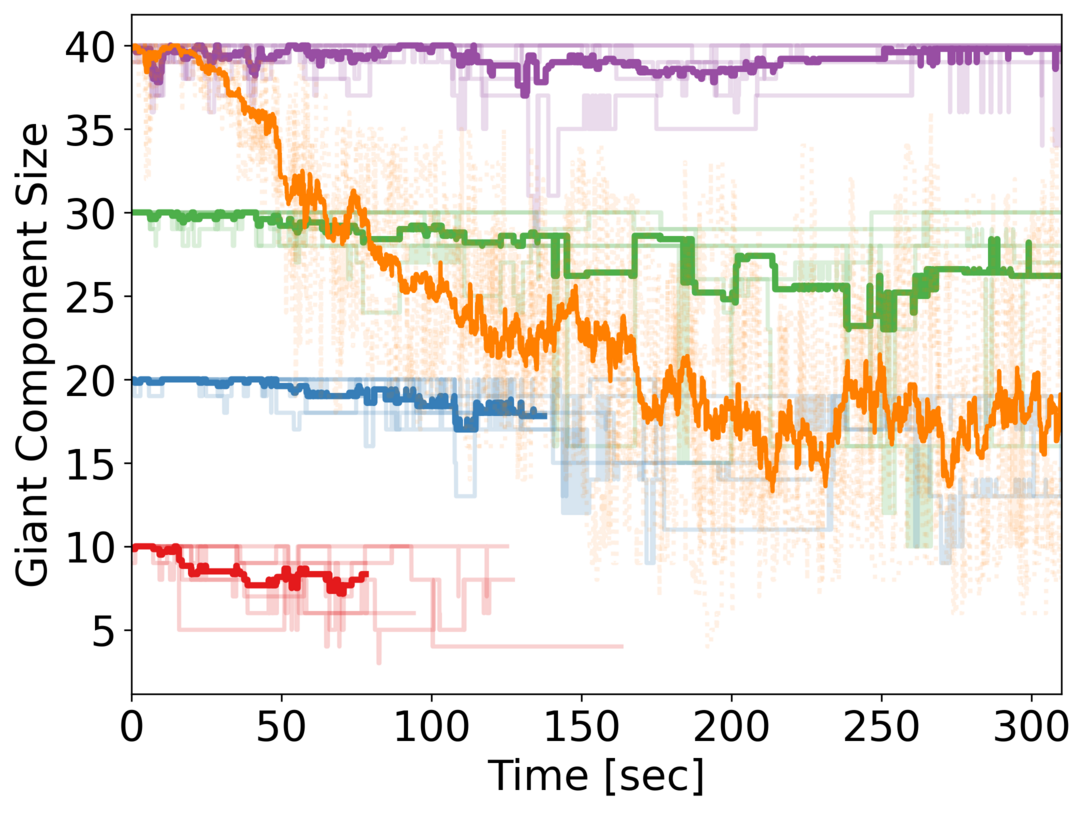}}%
    \hfill
    \subcaptionbox{}{\includegraphics[height=\figTwoWidth in]{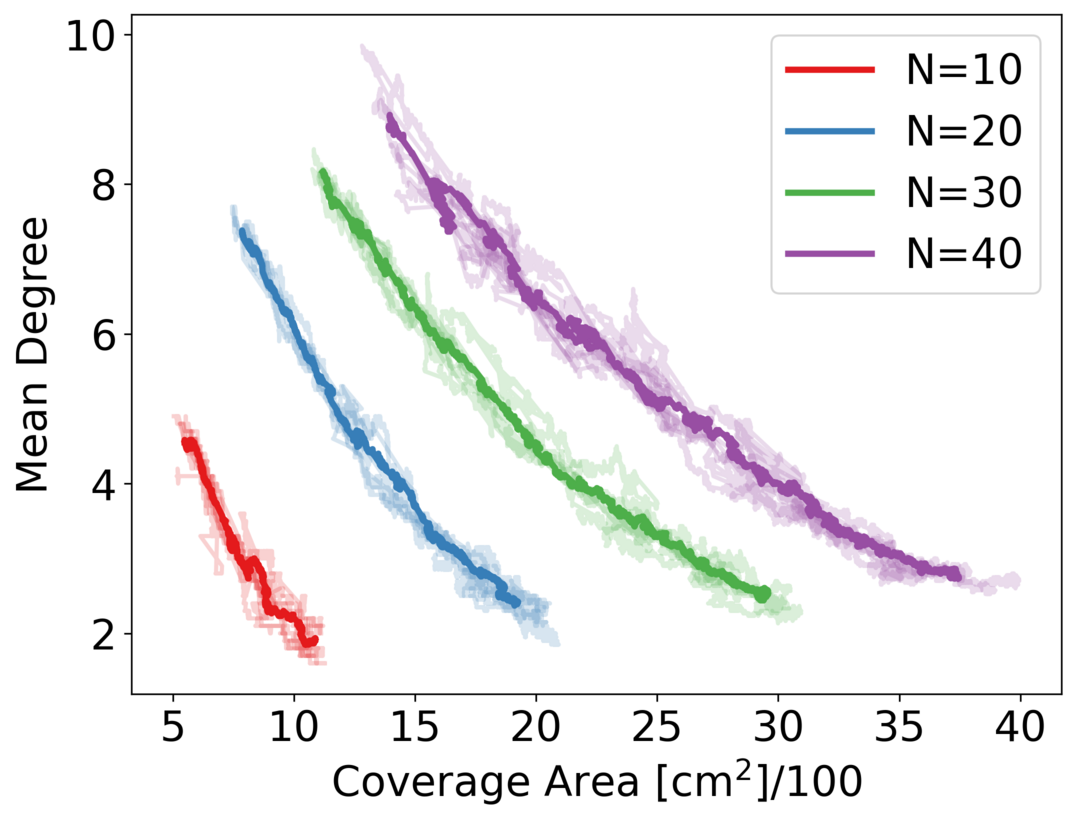}}%
\caption{Real-world experiments of the dispersion algorithm while preserving the network connectivity for 5 repetitions and swarm sizes of \{10,20,30,40\}: a)~covered area by the swarm, b)~mean degree of the communication network, c)~number of nodes in the giant component of the network versus time. The orange line shows the result for a diffusion algorithm without preserving the connectivity of the network for N$=40$ robots. d)~Trade-off between network connectivity and coverage area. The transparent lines show individual experiment, and the solid lines are the mean value for a corresponding swarm size. The mean value truncated as soon as the shortest experiment finishes.}
\label{fig:Res_Dispersion_ArCov_mDeg_gntComp}
\end{figure*}
\subsection{Exploration}
\label{subsec:exploration}
{\color{newChanges}With exploration, the variation and diversity of information available to the swarm increases.} 
During the exploration phase, no information is aggregated. As we demonstrated before~\cite{raoufi2021speed}, the exploration phase reduces the trueness error (systematic bias). In principle, any dispersion behavior may achieve the goal {\color{newChanges}in an unbounded environment}. 
However, due to limited connectivity of the distributed robots, a \emph{pure} random dispersion may disconnect robots from their neighbors and fragment the network.  Blind random motion in an unbounded environment is dangerous as robots might get lost and never find their way back to the swarm~{\color{newChanges}\cite{hornischer2021cimax}}.
\par
As an alternative, we suggest a random walk while preserving the connectivity of the network. A~robot requires to know the approximate distance to its neighbors. We will show that even with noisy distance estimations the method is able to {\color{newChanges}keep the swarm largely connected}. With Kilobots, the estimation of the distance is calculated by considering the strength of the received infra-red (IR) signal~\cite{rubenstein2012kilobot}. Hence, making the random walk conditional on the distance to the nearest neighbor is the algorithm we implemented on robots. 
{\color{newChanges}Once the minimum distance to local neighbors goes below the threshold, the robot stops and waits for its local neighbors to finish their random walk, then it switches to exploitation. Violations of the desired distance take the robot back to dispersion phase.}
By the end of this phase, the collective has the potential to make a less biased (or bias-free) estimation. Then, the swarm exploits the information distributed throughout the collective to increase the precision. See Fig.~\ref{fig:ATP_err__XprXpt} for an illustration on how exploration and exploitation can modulate the trueness and precision components of the total accuracy error.
\subsection{Exploitation}
\label{subsect:exploitation}
Exploitation operates not only in the information domain, but also in the real physical space. By exploiting the information contained in the swarm, the collective estimation converges to the mean value in the information domain. The exploitation in the physical space results in individual robots converging towards the mean contours of the environmental field. Here, we introduce two mechanisms for each of the domains: local averaging and consensus-based phototaxis.
\subsubsection{Local averaging}
The first part of exploitation is used to {\color{newChanges}reach} consensus in the information domain, which is achieved by local communication of robots. The results of interactions in this phase facilitate the wisdom of crowds effect~\cite{simons2004many, surowiecki2005wisdom}, by enabling the agents to average their imperfect estimates of environmental cues~\cite{hills2015exploration, becker2017network}. The updating rule comes from the local averaging of DeGroot model~\cite{degroot1974reaching}, and we modified it by adding a memory term~\cite{raoufi2021speed}. The ultimate updating rule is formulated as:
\begin{align}
& \hat{z}_{i}^{t+1} = \alpha  \hat{z}_{i}^{t} + \frac{1-\alpha}{1 + N_i}  s_{i}^{t} + \frac{1-\alpha}{1 + N_i} \sum\limits_{j \in \boldsymbol{N_i}} {\hat{z}_{j}^{t}}\ .
\label{Eq:consensus}
\end{align}
Here, each robot updates its estimation ($\hat{z}_{i}^{t+1}$) based on what it measures ($s_{i}^{t}$), and the average of its {\color{newChanges}$N_i$} neighbors' estimation, with a weighting factor $\alpha$.
Robots repeat these updates for a fixed number of iterations $t_\text{comm.}=100$. The output of this phase is the consensus value (although all robots might not have exactly the same opinion about the consensus.) Robots use this value as input for the next {\color{newChanges}phase}. 
\par
The updating equation (Eq.~\ref{Eq:consensus}) can be reformulated from a network point of view~\cite{olfati2007consensus, golub2010naive}. This would convert the model to a linear system whose transition matrix is the normalized weighted adjacency matrix of the network, its states are agents' estimation and the measurements are the inputs. Assuming the general system without input, the result of such local averaging, given that the network of communication is fully connected, is the mean value of information available within the collective~\cite{golub2010naive}. Later, we briefly discuss how the connectivity of the network (mean node degree, in particular) changes the dynamic of this system. 
%
%
%
%
\subsubsection{Consensus-based Phototaxis (CBPT)}
We implement a sample-based pseudo gradient descent for the motion of robots which implements homophily on networks. Homophily is the tendency to interact more with like-minded agents in a social group~\cite{khanam2022homophily}. We require a collective motion that moves robots sharing similar opinions closer to each other and thus establishes links~\cite{raoufi2021speed}. As a pseudo gradient descent method, we choose the bio-inspired phototaxis method. By CBPT the robots are guided to areas where light measurements match the consensus value. 
%
%
%
\section{Metrics and Setup}
\subsection{Covered Area}
We measure the area that is covered by the swarm. We consider a disk centered at each robot's position with radius ($r_\text{cover}$). For Kilobots, we choose $r_\text{cover} = 3r_\text{rob} = 5\ \text{cm}$ which is roughly half its communication range {\color{newChanges}($r_\text{rob}$~is robot radius)}. We calculate the collective coverage as the (non-overlapping) intersection of  areas~$A_{\text{cover},i}^{(x_i,y_i)}$ with $\|A_{\text{cover},i}^{(x_i,y_i)}\|=\pi r_\text{cover}^2$ covered by each robot~$i$ {\color{newChanges}located} at~$(x_i,y_i)$:
\begin{equation}
    A_\text{cover} = \bigcap\limits_{i=1}^{\text{N}}A_{\text{cover},i}^{(x_i,y_i)} 
    \;.
\end{equation}
\subsection{Network Properties}
The inter-agent communication network plays a critical role for the whole scenario. It is challenging to determine the existence of actual robot-robot communication links forming the network, as these links are noisy and difficult to extract from the robot swarm during an experiment. For simplicity we assume that if the distance between two robots is less than the average communication range, then there is a link. The communication range is assumed to be $r_\text{comm}=10\ \text{cm}$~\cite{pinciroli2018simulating}. The links are estimated based on robot positions and distances obtained from tracking via a top-view camera. False positives and negatives for links between robots are possible as this is only an estimation.
\par
We record the connectivity of the network by measuring the connectivity using two metrics: mean node degree and giant component size. Although the communication network of Kilobots is not necessarily undirected (signal strength is not always symmetric), we assume an undirected network for simplicity. In- and out-degree of all nodes are equal as well as the \emph{mean} in- and out-degree. As second network metric we use giant component sizes, that is the number of nodes in the largest  connected component of the network. This way we quantify how many robots have disconnected from the main cluster (implemented with NetworkX Library~\cite{hagberg2008exploring}).
\par
\subsection{Accuracy Metrics}
%
%
Collective estimation (accuracy) error is decomposed into trueness and precision error, which relates to the bias and variance decomposition of the total error. {\color{newChanges}We showed that the generality and case-independence of these metrics enable their usage in various conditions (see~\cite{raoufi2021speed} for details).} We assume as ground truth for estimation the mean value of the light intensity across the environment ${z}_\text{gt}=\bar{z}_\text{env}$. By defining the individual estimation for robot~$i$ as $\hat{z}_i$ and {\color{newChanges} collective estimation as $\hat{z}_\text{col}=\sum_{i=1}^{N}\hat{z}_i / {N}$}, we obtain for trueness, precision, and accuracy errors:
\begin{align}
    E_\text{T} =& (\hat{z}_\text{col} - {z}_\text{gt})^2 \ , 
    E_\text{P} = \frac{1}{\text{N}} \sum\limits_{i=1}^\text{N}(\hat{z}_i - \hat{z}_\text{col})^2 \ , \\
    E_\text{A} =& \frac{1}{\text{N}} \sum\limits_{i=1}^\text{N}(\hat{z}_i - {z}_\text{gt})^2 = E_\text{T} + E_\text{P} \ . 
\end{align}
As we have no direct access to a robot's current estimation, we use its position as an indicator of its estimation. For each environmental distribution, there is a mapping between the camera-detected Cartesian robot positions and the coordination of interest. For example, in the radial distribution of Fig.~\ref{fig:snapShots_pl_ATP_err_XprXpt}, the mapping $m(x_i,y_i)$ is:
\begin{equation}
    \hat{z}_i = r_i = m(x_i,y_i) = \sqrt{(x_i-x_c)^2 + (y_i-y_c)^2},
\end{equation}
where, $(x_c,y_c)$ is the distribution's center, and $(x_i,y_i)$ is the detected robot's position in the captured frame.
\\
\subsection{Experimental Setup}
In our experiments we use Kilobot robot swarms~\cite{rubenstein2012kilobot} of up to 40 robots, on a {\color{newChanges}$90\times90\ \text{cm}^2$ arena of a $1.5\times2.5\ \text{m}^2$ white-board}. For tracking we use a downward-facing camera and Hough circle transformation from OpenCV Library~\cite{opencv_library}. {\color{newChanges} Otherwise mentioned, we used the same parameters as~\cite{raoufi2021speed}.} 
\newcommand\figTwoHeight{0.8}
\begin{figure}[b]
\centering
    \subcaptionbox{}{\includegraphics[height=\figTwoHeight in]{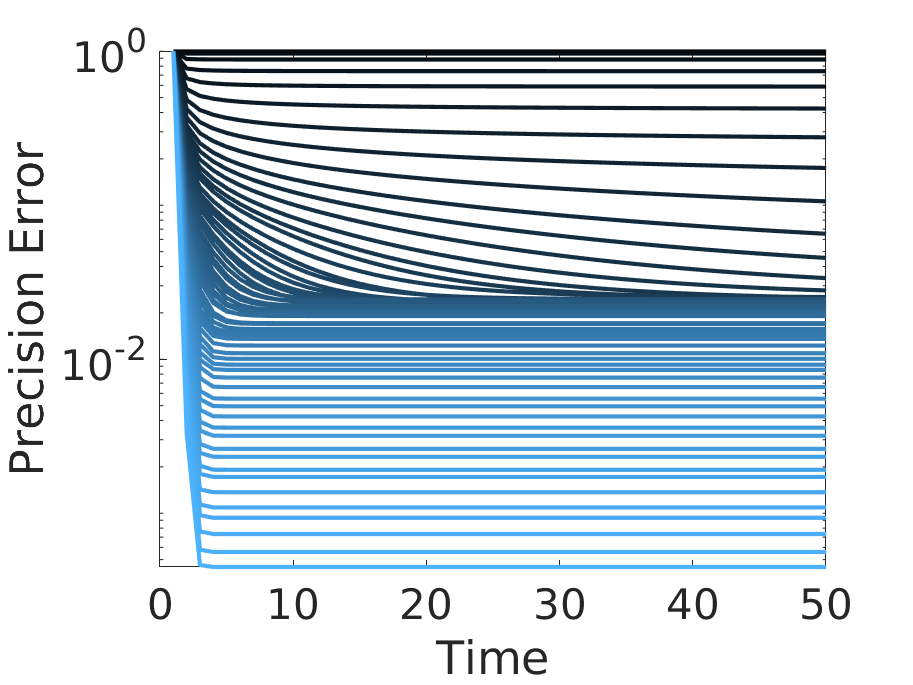}}%
    \hfill
    \subcaptionbox{}{\includegraphics[height=\figTwoHeight in]{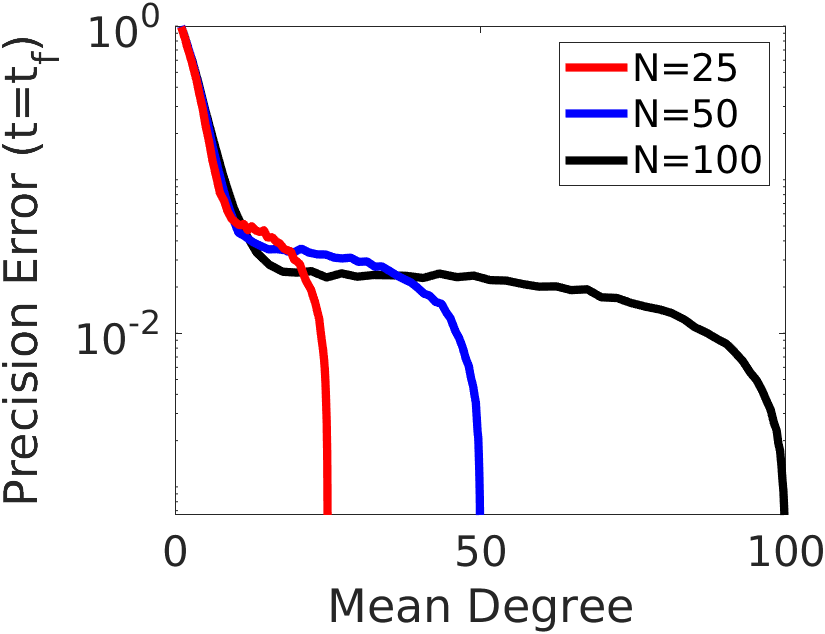}}%
    \hfill
    \subcaptionbox{}{\includegraphics[height=\figTwoHeight in]{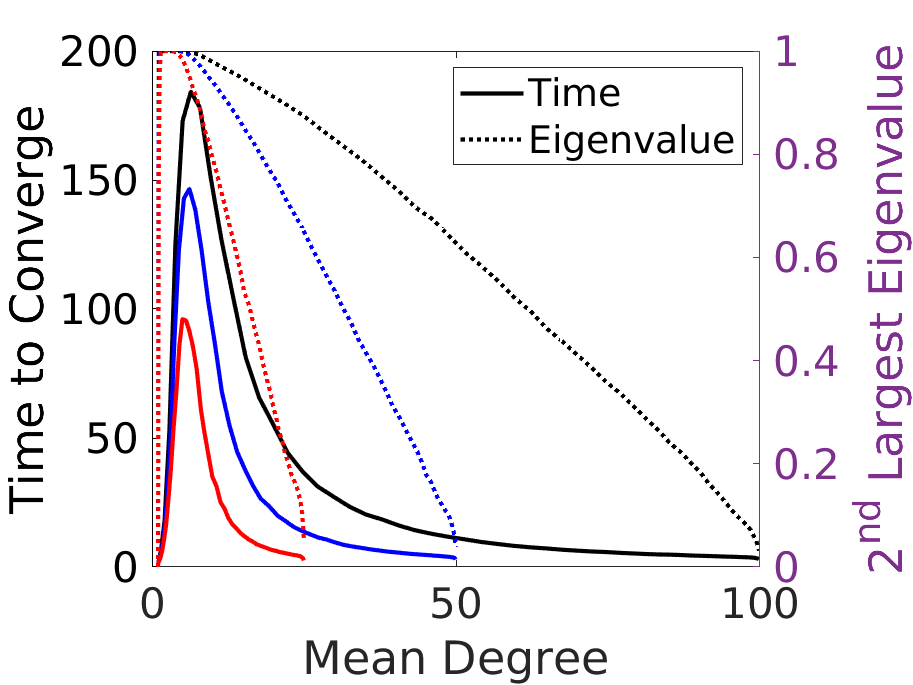}}%
\caption{The simulation of consensus model on static networks with different connectivity. a) Time evolution of precision error for different networks (darker lines indicate lower connectivity). b) Steady-state precision error (last time step) versus mean node degree, c)~time to achieve a steady state (solid lines) and second largest eigenvalue of adjacency matrix (dotted lines) versus mean node degree. The results are the average of 1000 {\color{newChanges}independent Monte Carlo simulations.}}
\label{fig:consensus}
\end{figure}
%
%
%
%
\section{Results and Discussion}
\newcommand{\figThreeHeight}{1.15}
\begin{figure*}[ht]
\centering
    \subcaptionbox{}{\includegraphics[height=\figThreeHeight in]{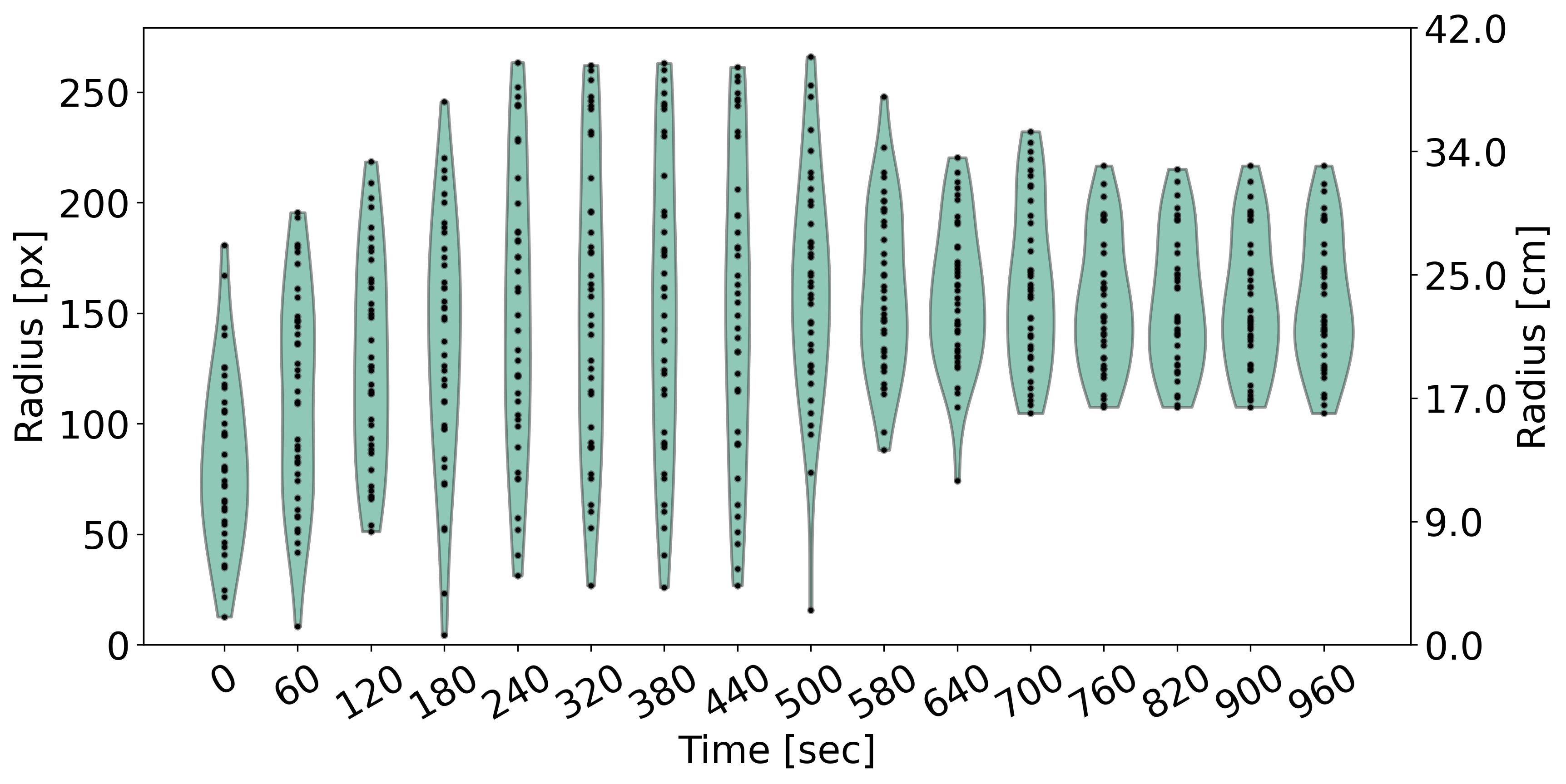}}%
    \hfill
    \subcaptionbox{}{\includegraphics[height=\figThreeHeight in]{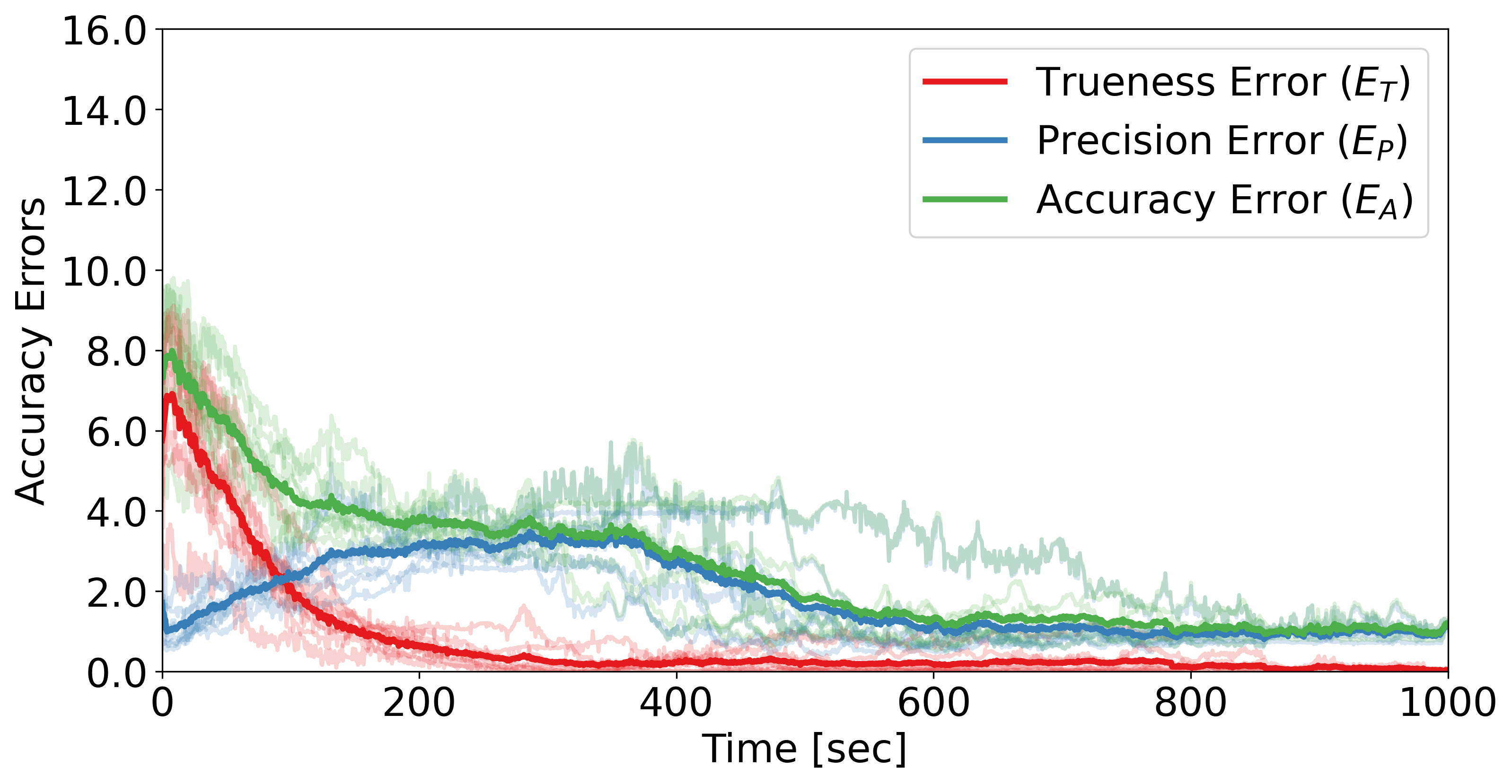}}%
    \hfill
    \subcaptionbox{}{\includegraphics[height=\figThreeHeight in]{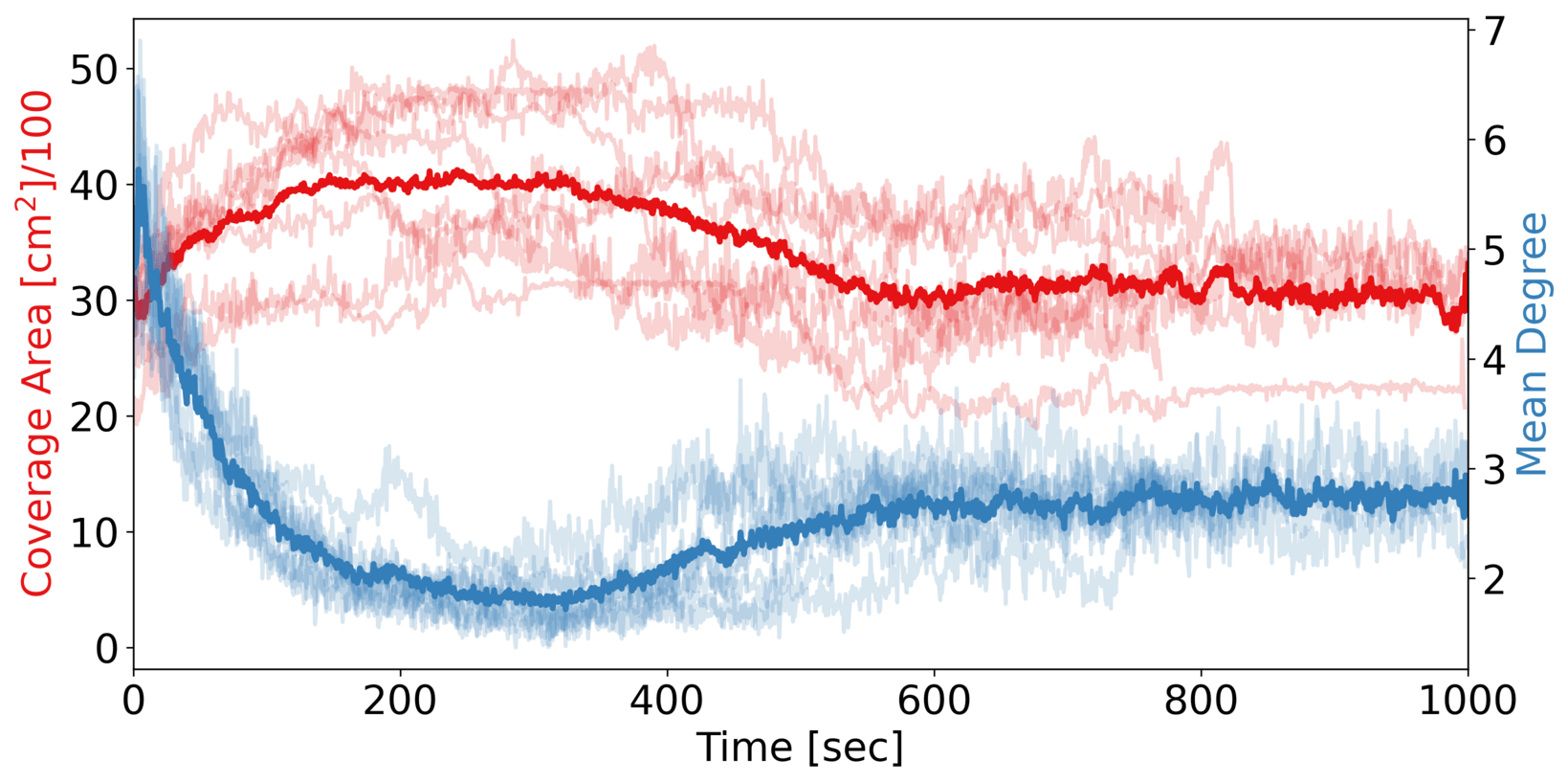}}
\caption{Real-world experiments of the full scenario in an environment with radial distribution for $N=40$ Kilobots. a) The distribution of robots of a single experiment in the {\color{newChanges}polar coordinate system}, b) the accuracy errors, c) the coverage area and mean node degree of the network over time. The transparent lines show the result of 8 independent real robot experiments, and the solid lines are the average over different experiments.}
\label{fig:Res_contCapturing}
\end{figure*}
We study each component of our scenario (dispersion, consensus, CBPT) as stand-alone swarm tasks. Later, we combine these components to form a complex scenario. 
\subsection{Dispersion}
\label{sec:dispersion}
The aim of dispersion is to increase covered area. We measure how much area is covered by robots (Fig.~\ref{fig:Res_Dispersion_ArCov_mDeg_gntComp}-a). To indicate the dynamic network structure, we measure the mean degree of the network (Fig.~\ref{fig:Res_Dispersion_ArCov_mDeg_gntComp}-b). The results in Fig.~\ref{fig:Res_Dispersion_ArCov_mDeg_gntComp}, indicate that initially the collective starts from a dense distribution with low coverage area and high connectivity in the network. Due to dispersion, the collective expands and covers larger area while the mean degree decreases. This increase in the covered area can lead to a lower trueness error in the collective estimation. 
%
%
The network gets sparser (reduced node degrees) {\color{newChanges} while the giant component size of the network does not change significantly, suggesting that the network connectivity is largely preserved.} 
Later we show how reduced connectivity results in lower speed of convergence during the decision-making process. Both the covered area and mean degree converge to steady state values. Once robots stop moving, we finish the experiment. 
\par
In Fig.~\ref{fig:Res_Dispersion_ArCov_mDeg_gntComp}-c, we show the size of the giant component. The algorithm keeps the majority of the swarm connected while a few robots disconnect from the swarm. In our analysis we found that often two (or more) robots stick to each other and while measuring strong signals from each other, they continue moving. They detach from the swarm, although they are members of a small cluster. As a control experiment, we tested a random walk diffusion algorithm that does not try to preserve connectivity (solid orange line in Fig.~\ref{fig:Res_Dispersion_ArCov_mDeg_gntComp}-c). Almost half of the swarm disconnects within three minutes. In comparison, our algorithm preserves connectivity well.
\subsection{Consensus}
Consensus occurs only in the information domain, which makes it difficult to measure in a real robot experiment. However, we simulated the consensus algorithm on a static network in order to show how the precision error changes over time (Fig.~\ref{fig:consensus}-a) and how its dynamics change with changing network properties, namely mean degree. We studied spatial networks with N$=\{25,50,100\}$ nodes and different connectivity to investigate the role of mean degree. For doing so, we distributed N agents uniformly in an environment, and drew a deterministic network with a specific communication range. Then, we varied the communication range (ratio to the environment size) to achieve networks with various mean degree. As agents share and update their estimation about the mean value of the distribution, they converge to the consensus estimation, and thus the precision error decreases (Fig.~\ref{fig:consensus}-a)--this is the well-known speed-vs-accuracy trade-off happening over the course of decision-making. 
\par
In Fig.~\ref{fig:consensus}-b, we show how the mean node degree of the network influences the accuracy (precision) of the steady-state collective estimation. A higher mean degree leads to a lower precision error. With respect to speed of consensus, we measured the time to reach a steady state using a threshold ($\delta=1e-4$) and recorded the first passage time of the precision error. The peaks in Fig.~\ref{fig:consensus}-c show the slowest convergence time for a specific mean degree. The speed reduces significantly for lower and higher degrees. 
A~low or zero mean degree means there are few or no links in the network. Convergence is fast without information flow but not accurate.
As known from graph theory, the network is immediately fully connected once the mean degree exceeds a critical value. This is where the second largest eigenvalue of the network adjacency matrix {\color{newChanges}becomes less than} one.
\newcommand{\figFourHeight}{0.95}
\begin{figure*}[!ht]
\centering
    \subcaptionbox{}{\includegraphics[height=\figFourHeight in]{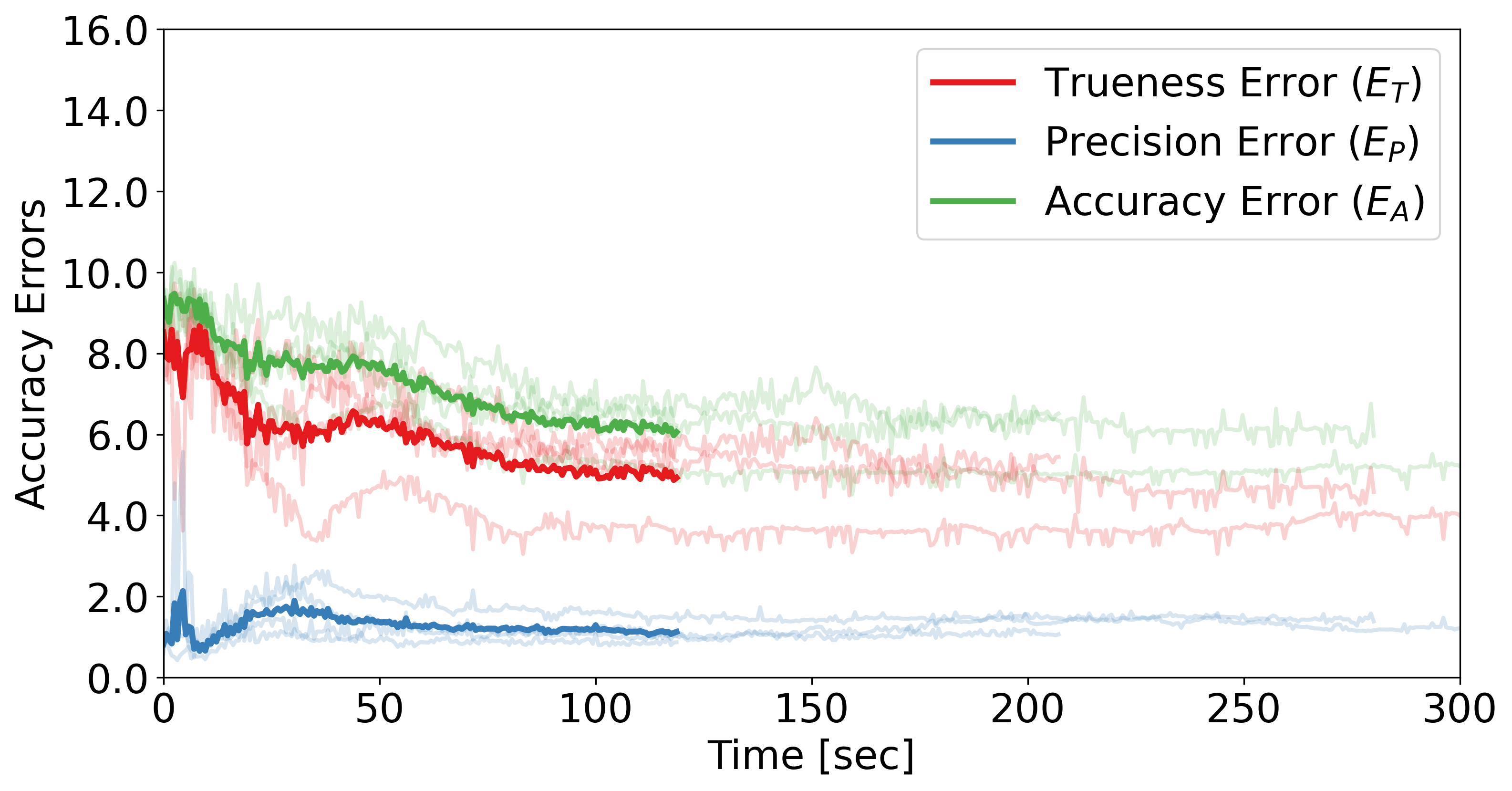}}%
    \hfill
    \subcaptionbox{}{\includegraphics[height=\figFourHeight in]{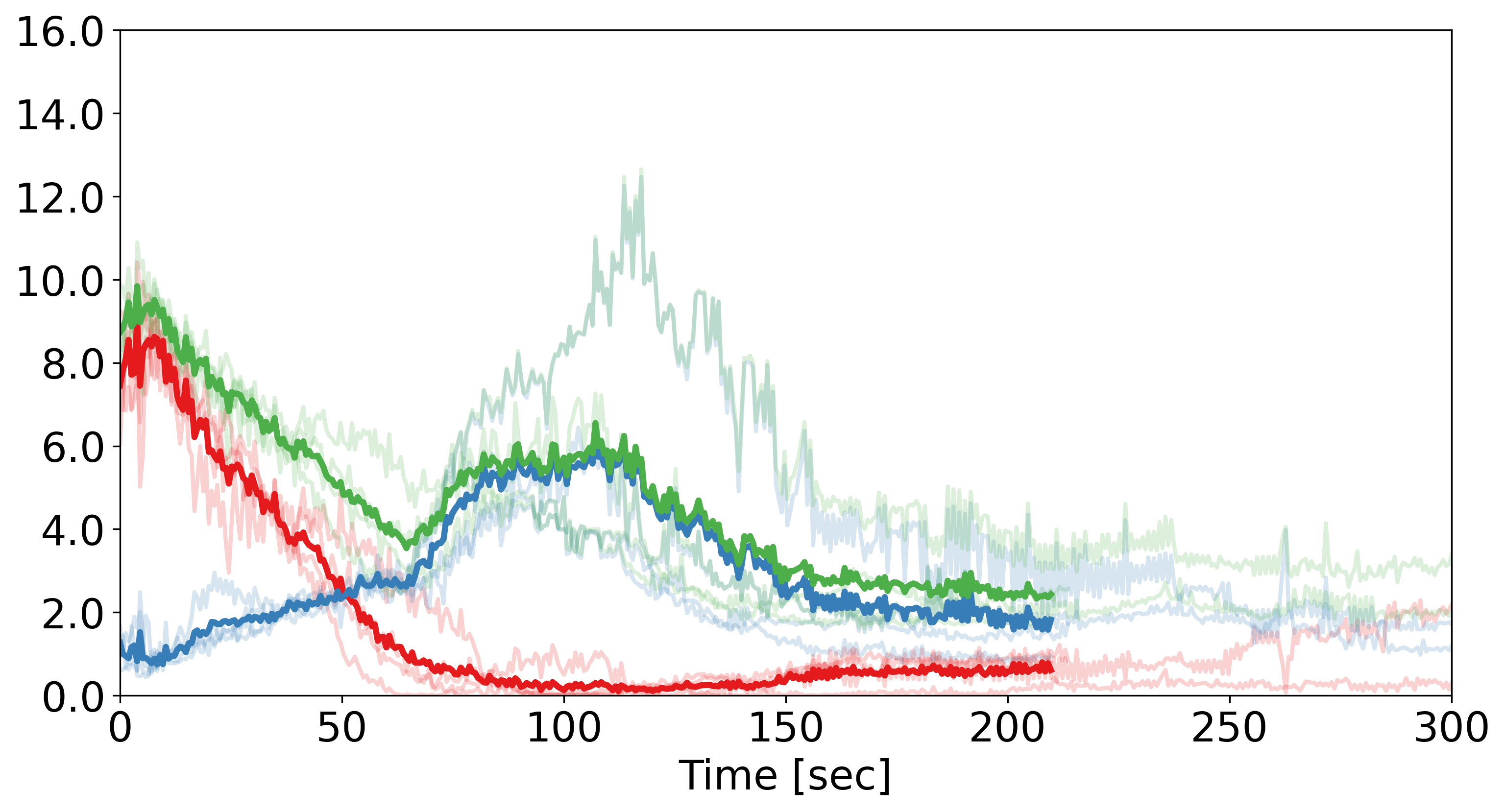}}%
    \hfill
    \subcaptionbox{}{\includegraphics[height=\figFourHeight in]{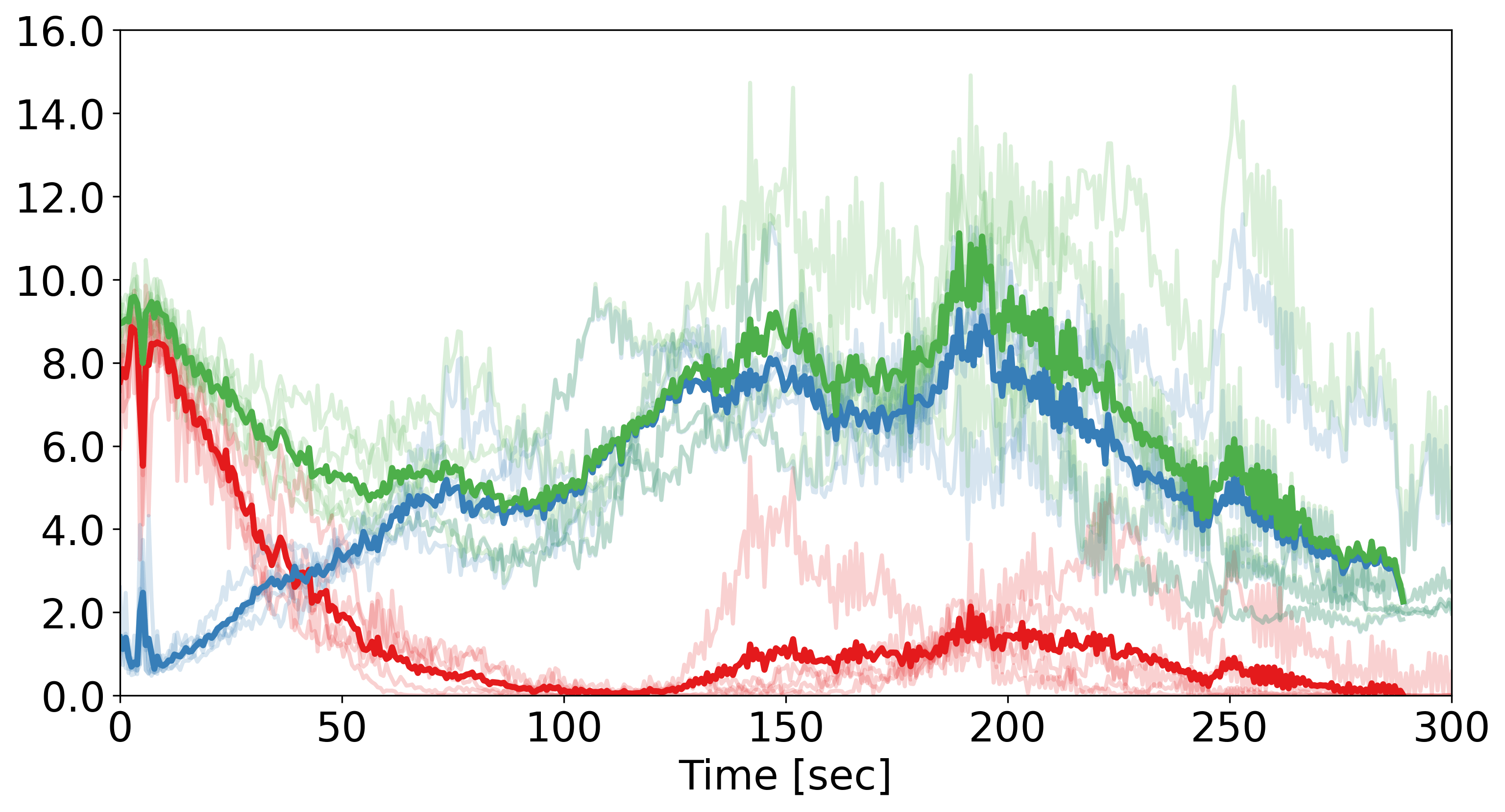}}%
    \hfill
    \subcaptionbox{}{\includegraphics[height=\figFourHeight in]{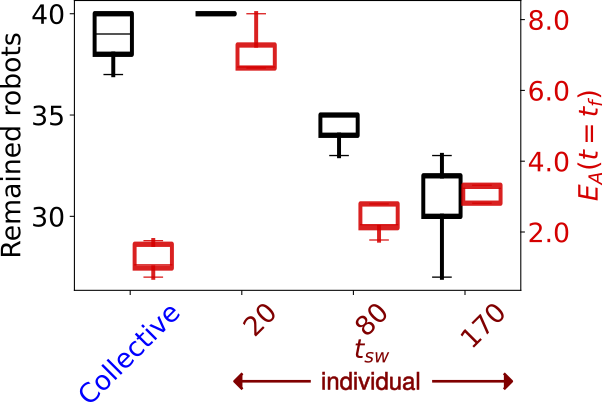}}
\caption{Real-world control experiments with $N=40$ Kilobots. The accuracy errors over time for the control experiments show the performance of the individualistic method. 
Each plot is the average of 5 real robot experiments for different switching times, a) $t_\text{sw} = 20$, b) $t_\text{sw} = 80$, c) $t_\text{sw} = 170$. d) Box plot of the number of robots remained within the area of interest at the last snapshot (black boxes) and the final accuracy error {\color{newChanges}(red boxes)} comparing the collective scenario and the control experiment (individual) with different switching times.}
\label{fig:Res_contCapturing_controlExp}
\end{figure*}
\subsection{Contour Capturing}
Next, we present our results for the scenario of contour capturing with a swarm of $N=40$ Kilobots. The  objective is to gather the robots at the contours with mean light distribution. First, we give results of our fully distributed collective method. Second, we define a control experiment without robot-robot communication as baseline for comparison.
\subsubsection{Collective Scenario--radial distribution}
Here we present our main result in real-world experiments with Kilobots for the whole scenario by assembling the above components: dispersion while keeping the network connected, local averaging to achieve consensus, and homophily by CBPT to approach the consensus value. 
For a radial light distribution, 
Fig.~\ref{fig:Res_contCapturing}-a shows the radial distribution of robot positions during the experiment. Initially, the robots are distributed rather densely close to the center ($r_i \rightarrow 0$). During the dispersion the distribution becomes more uniform by spreading to larger radii.
Then the local consensus finding with minimal movement starts while the spatial distribution of robots remains largely unchanged ($200<t<400$). In a third phase robots approach the mean contour line by CBPT and the distribution contracts around 160 pixels {\color{newChanges}($\approx 25\text{cm}$)}.
\par
The temporal evolution of the trueness, precision and accuracy errors is illustrated in Fig.~\ref{fig:Res_contCapturing}-b. The trueness error quickly drops to a small value by the end of the dispersion phase ($t\approx200$ sec). However, the variation is still large although the mean value of the radial distribution is close to the ground truth. Thus, in contrast to the accurate mean value of the collective, each robots' estimation is not yet accurate. This is because robots did not aggregate any information during dispersion.
But, now that the collective is less-biased, and the network is connected robots exploit the information available within the entire collective.
This is implemented via the local average from the consensus method (see eq.~\ref{Eq:consensus}). At time~$t\approx400$~s, the swarm arrives at a consensus on the information domain, but robot positions are still off the mean contour line. During the CBPT phase, robots approach the mean value in space and precision error is reduced. We observe both a low precision error and a low accuracy error. These results confirm our previous work in simulations~\cite{raoufi2021speed}. 
\par
The mean degree and area coverage of the swarm evolve in an anti-correlated manner. During dispersion, the swarm spreads out to cover more area and the spatial distribution gets sparser, hence, reducing the mean node degree.
But the process inverts during exploitation as robots get closer to each other and increase network connectivity. Covered area decreases because robots form a denser distribution around the contour line and the overlap area increases.
\subsubsection{Control experiment--no communication}
As control experiment, the robots do contour capturing without collaboration between robots or exchange of any information. During  exploration, each robot walks randomly while updating and aggregating its mean value estimation. Robots {\color{newChanges}iteratively} average over measured samples. The random walk is random diffusion and without effects by other robots (in difference to Sec.~\ref{sec:dispersion}). It stops after a predetermined number of samples ($t_\text{sw}$). Then robots switch to exploitation and follow the CBPT algorithm to approach the estimated mean light spot. We used three different switching times: $t_\text{sw} = \{20, 80, 170\}$.
\par
As seen in Fig.~\ref{fig:Res_contCapturing_controlExp}-a, a too short exploration ($t_\text{sw} = 20$) does only insufficiently reduce the trueness error (red line). Whereas the precision error (blue line) remains as high as the initial value due to insufficient spatial dispersal of robots.
In Fig.~\ref{fig:Res_contCapturing_controlExp}-b, a sufficiently long exploration ($t_\text{sw} = 80$) reduces the trueness error, and manages the temporarily high precision error ($t\approx100$~s).
Fig.~\ref{fig:Res_contCapturing_controlExp}-c indicates a too long
exploration phase resulting in a larger precision error. In our previous work~\cite{raoufi2021speed}, we already showed that (in a bounded environment) too late switching can cause the precision error to remain high  (for a limited time budget).
\\
The \emph{unbounded} environment is challenging as the swarm tends to loose more and more robots (lost connectivity) with increased exploration time (Fig.~\ref{fig:Res_contCapturing_controlExp}-d). In addition to the known speed-vs-accuracy trade-off, we find this new trade-off in unbounded environments. With uncontrolled diffusion, one does not only pay in speed for accuracy, but also in the number of robots that get lost.
\subsubsection{Collective Scenario--V-shape ramp distribution}
In the model simulations presented in~\cite{raoufi2021speed}, we showed that the algorithm is able to capture the mean contour line for different environmental distributions, including uni- and multi-modal ones. In this part, we tested another distribution that is of an inverted V-shape, with a peak on its diagonal as in Fig.~\ref{fig:Res_contCapturing_rotRamp}-a. The evolution of the distribution of robots over time (Fig.~\ref{fig:Res_contCapturing_rotRamp}-b) demonstrates how the swarm expands uniformly up until the exploitation phase. Then, they branch into two different clusters; one on the top left and the other on the bottom right of the diagonal. The accuracy errors of Fig.~\ref{fig:Res_contCapturing_rotRamp}-(c) have the same qualitative trends as in Fig.~\ref{fig:Res_contCapturing} for radial distribution. However, the remaining precision error at the end of the experiments indicates that the problem here is more difficult to solve. We note that here the precision error represents the dominant contribution to the total error. 
\renewcommand{\figTwoHeight}{1.15}
\begin{figure}[hb]
\centering
    \subcaptionbox{}{\includegraphics[height=\figTwoHeight in]{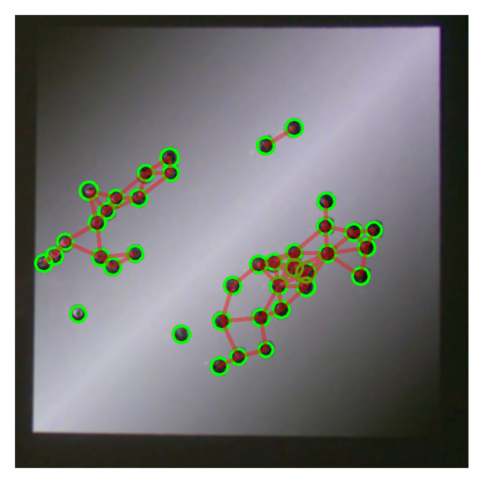}}%
     \hfill
    \subcaptionbox{}{\includegraphics[height=\figTwoHeight in]{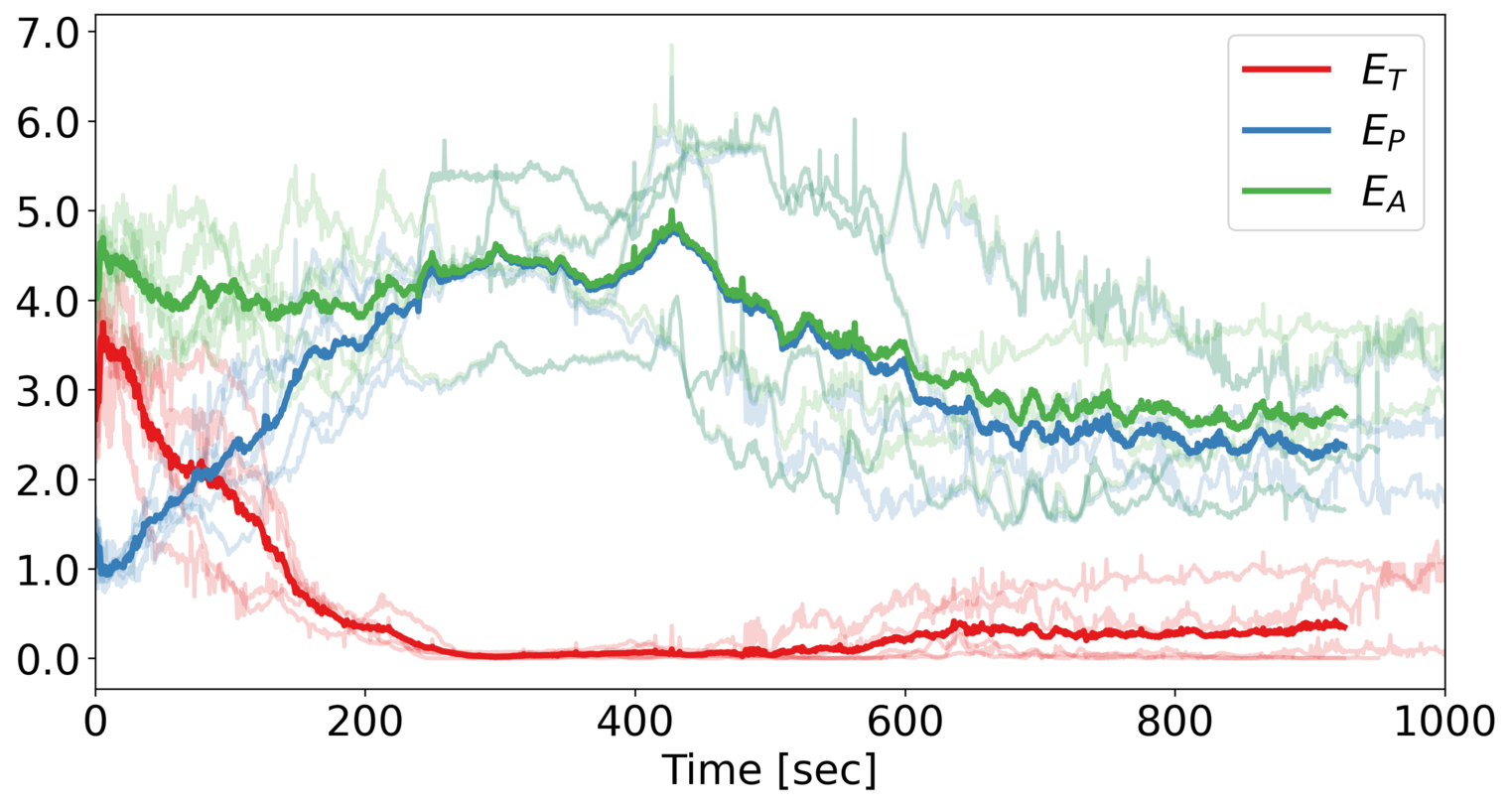}}
\caption{Real-world results for the scenario with diagonal distribution: a)~representative example of final robot distribution showing the position of two major clusters on each side of the ramp. b)~Three errors types over time (6~repetitions).}
\label{fig:Res_contCapturing_rotRamp}
\end{figure}
%
%
%
%
\section{Conclusion}
Starting from our previous work on the speed-accuracy trade-off in collective estimation~\cite{raoufi2021speed}, we have successfully implemented a real robot swarm (Kilobots) to capture a contour in a continuous environmental field in an unbounded arena.  
Our dispersion method {\color{newChanges}largely} preserves connectivity of the swarm and minimizes losing robots {\color{newChanges} during exploration}. 
%
As another component, we introduced a sample-based optimization method inspired by phototaxis that makes the Kilobots approach the desired contour. 
We added a light conductor to the robot (minimizing shadows on the sensor) to improve light measurements. 
This seems to be a novel implementation of a gradient ascent for Kilobots with various potential applications. 
{\color{newChanges} The codes we used in this paper are available on GitHub \cite{MRaoufi_Github}.}
\par
Previously we showed that besides the {\color{newChanges} speed-vs-accuracy} there are also  exploration-vs-exploitation trade-offs~\cite{raoufi2021speed} that are generally non-trivial to resolve. With our new dispersion method, an optimal switching time to finish exploration is not required anymore. The swarm automatically ends dispersion at supposed best achievement constrained by connectivity. Here we discussed another trade-off induced by dynamic network topologies. During exploration, the temporarily low mean degree slows down collective decision-making. But the swarm expansion improves the accuracy of the estimation. 
\par
In future work, we plan to study contour-capturing scenarios in dynamic environments. We also plan to analyze scalability and test different light distributions. 
\section*{Acknowledgment}
We thank Marshall Lutz Mykietyshyn and Noran Abdelsalam for their contribution to real robot experiments. 
%
%
%
%
\bibliographystyle{IEEEtran}
\bibliography{refs}
\end{document}